arXiv:1604.05144v1 [cs.CV] 18 Apr 2016

# ScribbleSup: Scribble-Supervised Convolutional Networks for Semantic Segmentation

Di Lin[1*] Jifeng Dai[2] Jiaya Jia[1] Kaiming He[2] Jian Sun[2]
[1]The Chinese Univeristy of Hong Kong [2]Microsoft Research

## Abstract

*Large-scale data is of crucial importance for learning semantic segmentation models, but annotating per-pixel masks is a tedious and inefficient procedure. We note that for the topic of interactive image segmentation, scribbles are very widely used in academic research and commercial software, and are recognized as one of the most user-friendly ways of interacting. In this paper, we propose to use scribbles to annotate images, and develop an algorithm to train convolutional networks for semantic segmentation supervised by scribbles. Our algorithm is based on a graphical model that jointly propagates information from scribbles to unmarked pixels and learns network parameters. We present competitive object semantic segmentation results on the PASCAL VOC dataset by using scribbles as annotations. Scribbles are also favored for annotating stuff (*e.g.*, water, sky, grass) that has no well-defined shape, and our method shows excellent results on the PASCAL-CONTEXT dataset thanks to extra inexpensive scribble annotations. Our scribble annotations on PASCAL VOC are available at* `http://research.microsoft.com/en-us/um/people/jifdai/downloads/scribble_sup`.

## 1. Introduction

Recent success of semantic segmentation lies on the end-to-end training of convolutional networks (*e.g.*, [21]) and large-scale segmentation annotations (*e.g.*, [18]). Even though semantic segmentation models are being improved rapidly, large-scale training data still have apparent benefits for accuracy, as evidenced in [5, 31, 24, 7, 20].

But it is painstaking to annotate precise, mask-level labels on large-scale image data. For example, powerful interactive tools [1] are adopted for annotating the MS COCO dataset [18], but it still takes minutes for an experienced annotator labeling one image [18]. The interactive tools of [1] require annotators to draw polygons along object boundaries, and the number of polygon vertexes in an image can be a few tens. This time-consuming task may limit the amount of data that have mask-level labels.

The procedure of annotating segmentation masks is very similar to *interactive image segmentation* [4], which has been a widely studied problem in the past decade [4, 26, 17, 11, 16, 19, 6]. Among various forms of user interactions, *scribbles* are particularly popular [4, 17, 11, 16, 19, 6] and have been recognized as a user-friendly way of interacting. A commercial implementation of scribble-based interactive segmentation is the *Quick Selection*[1] tool in Adobe Photoshop, which is a prevalent tool for selecting complicated regions in user photos.

Driven by the popularity of using scribbles for interactive image segmentation, we believe that it is more efficient to use scribbles to annotate images. By dragging the cursor in the center of the objects, the annotators need not carefully outline the object boundaries. It is also easier to use scribbles to annotate "stuff" (water, sky, grass, *etc.*) that may have ambiguous boundaries and no well-defined shape. Fig. 1 shows an example of scribble annotations.

An obvious way of using scribbles to ease annotating is to perform interactive image segmentation (*e.g.*, Quick Selection) when the annotators are interacting. But we argue that to obtain precise masks that play as ground truth, the annotators may still have to do many "touch-ups" on the imperfect outcome, which are inefficient. Instead of doing so, we consider to directly *use the sparse scribbles for training semantic segmentation models*. The annotators just need to provide a few scribbles on the regions which they feel confident and easy to draw.

With this scenario, in this paper we develop an algorithm that exploits scribble annotations to train convolutional networks for semantic segmentation. This problem belongs to the category of weakly-supervised learning, and occupies a middle ground between image-level supervision and box-level supervision. Comparing with image-level annotations [24], scribbles provide location information at a few pixels, which should lead to better results; comparing with box-level annotations [24, 7], scribbles are lack of determinate

*This work was done when Di Lin was an intern at Microsoft Research.

[1]`https://helpx.adobe.com/photoshop/using/making-quick-selections.html`

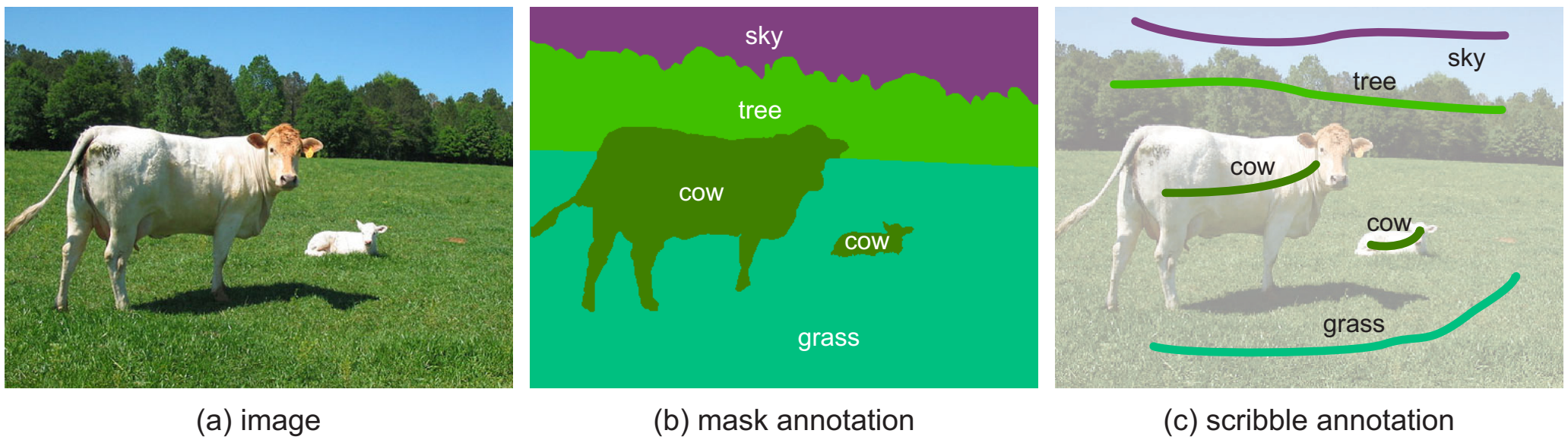

Figure 1. An image annotated with per-pixel masks (middle) and sparse scribbles (right). Scribbles are more user-friendly for annotators.

bounds of objects and so are more ambiguous.

We tackle scribble-supervised training by optimizing a graphical model. The graphical model propagates the information from the scribbles to the unmarked pixels, based on spatial constraints, appearance, and semantic content. Meanwhile, a fully convolutional network (FCN) [21] is learned, which is supervised by the propagated labels and in turn provides semantic predictions for the graphical model. We formulate this model as a unified loss function, and develop an alternating method to optimize it.

To evaluate our method, we use the Amazon Mechanical Turk platform to annotate scribbles on the PASCAL VOC datasets. Using training images with sparse scribble annotations, our method shows reasonable degradation compared to the strongly-supervised (by masks) counterpart on PASCAL VOC 2012. Compared with other weakly-supervised methods that are based on image-level or box-level annotations, our scribble-supervised method has higher accuracy. Furthermore, by exploiting inexpensive scribble annotations on PASCAL VOC 2007 (which has no mask annotations), our method achieves higher accuracy on the PASCAL-CONTEXT dataset (that involves objects and stuff) than previous methods that are not able to harness scribbles. These results suggest that in practice scribble annotations can be a cost-effective solution, and accuracy can be increased by larger amount of inexpensive training data. Our scribble annotations on PASCAL VOC are available at `http://research.microsoft.com/en-us/um/people/jifdai/downloads/scribble_sup`.

## 2. Scribble-Supervised Learning

An annotated scribble (*e.g.*, Fig. 1) is a set of pixels with a category label. The scribbles are sparsely provided, and the pixels that are not annotated are considered as unknown. Our training algorithm uses a set of images with annotated scribbles, and trains fully convolutional networks [21] for semantic segmentation.

We argue that scribble-based training is more challenging than previous box-based training [24, 7]. A box annotation can provide determinate bounds of the objects, but scribbles are most often labeled on the internal of the objects. In addition, box annotations imply that all pixels outside of the boxes are not of the concerned categories. This implication is not available for scribbles. In the case of scribbles, we need to propagate information from the scribbles to all other unknown pixels.

Our training algorithm accounts for two tasks. For the first task, our training algorithm propagates the semantic labels from the scribbles to other pixels and fully annotates the images; for the second task, our training algorithm learns a convolutional network for semantic segmentation. These two tasks are dependent on each other. We formulate the training as optimizing a unified loss function that has two terms. The dependency between the two tasks is made explicit in an alternating training solution.

### 2.1. Objective Functions

We use a graphical model to propagate information from scribbles to unknown pixels. We build a graph on the super-pixels of a training image. A vertex in the graph represents a super-pixel, and an edge in the graph represents the similarity between two super-pixels (see Fig. 2). We use the method of [10] to generate super-pixels.

We denote a training image as $X$, and its set of non-overlapping super-pixels as $\{x_i\}$ satisfying $\bigcup_i x_i = X$ and $x_i \bigcap x_j = \varnothing, \forall i, j$. The scribble annotations of this image are $S = \{s_k, c_k\}$ where $s_k$ is the pixels of a scribble $k$ and $0 \leq c_k \leq C$ is the scribble's category label (assuming there are $C$ categories and $c_k = 0$ for background). For a super-pixel $x_i$, we want to find a category label $0 \leq y_i \leq C$. The set of $\{y_i\}$ is denoted as $Y$. The labeling $Y$ provides full annotations of the image. Our objective function is:

$$\sum_i \psi_i(y_i|X, S) + \sum_{i,j} \psi_{ij}(y_i, y_j|X), \tag{1}$$

where $\psi_i$ is a *unary term* involving the super-pixel $x_i$, and $\psi_{ij}$ is a *pairwise term* involving a pair of super-pixels $x_i$ and $x_j$. Formulation in this form is widely used for interactive image segmentation [4, 26, 17, 11, 16, 19].

In our scenario, the unary term $\psi_i$ has two parts. The first part is based on scribbles and denoted as $\psi_i^{scr}$. We define

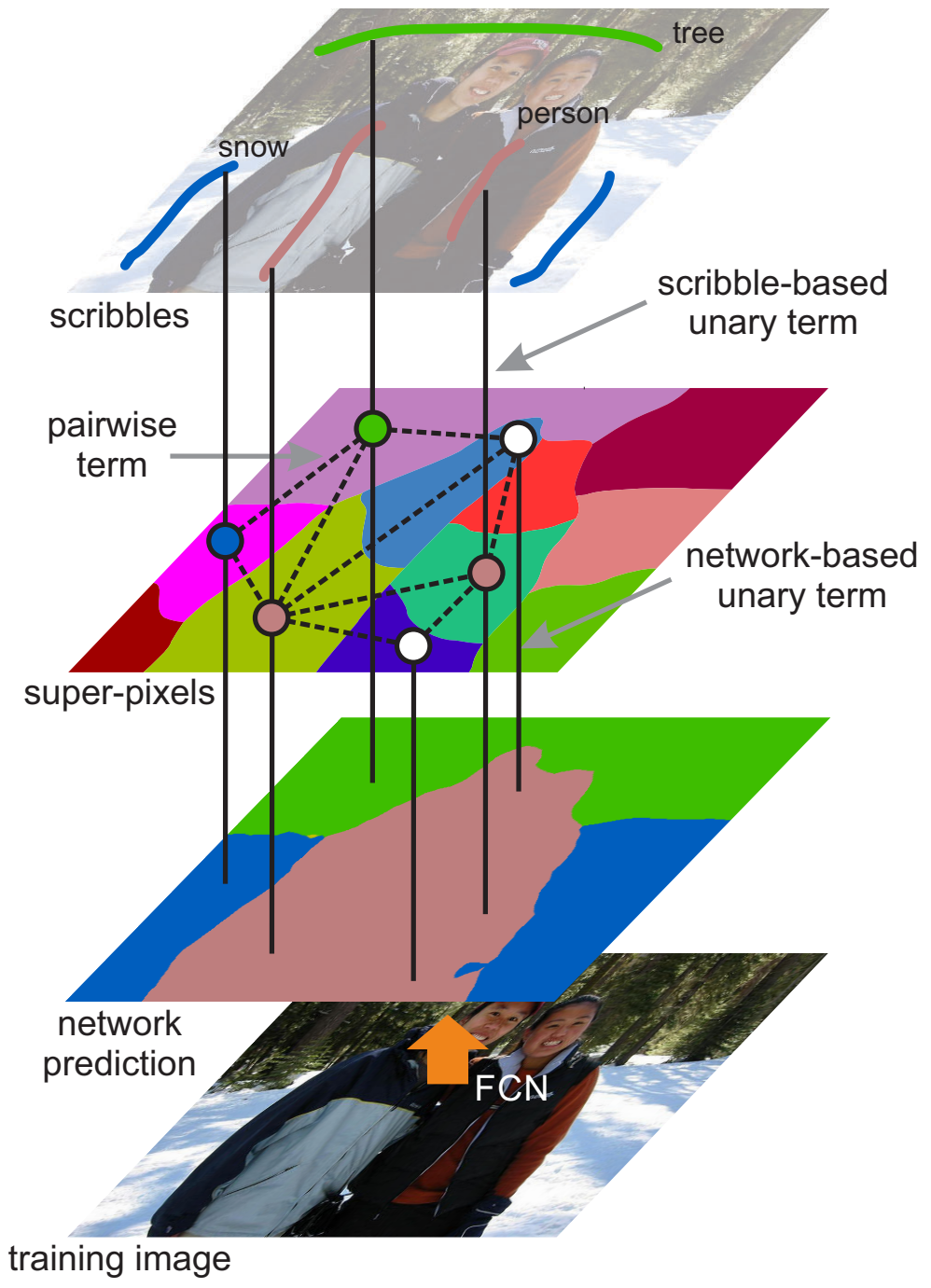

Figure 2. Overview of our graphical model. The graph is built on the super-pixels of the training image. The scribble-based unary term imposes constraints from the user annotations. The network-based unary term respects the predictions from a fully convolutional network. The pairwise terms are used to propagate information into the unmarked pixels. In this figure, the super-pixels are for the illustration purpose and are not their actual shapes.

this term as follows:

$$\psi_i^{scr}(y_i) = \begin{cases} 0 & \text{if } y_i = c_k \text{ and } x_i \bigcap s_k \neq \varnothing \\ -\log(\frac{1}{|\{c_k\}|}) & \text{if } y_i \in \{c_k\} \text{ and } x_i \bigcap S = \varnothing \\ \infty & \text{otherwise} \end{cases} \quad (2)$$

In this formulation, the first condition means that if a super-pixel $x_i$ overlaps with a scribble $s_k$, then it has zero cost when being assigned the label $c_k$. In the second condition, if a super-pixel $x_i$ does not overlap with any scribble, it can be assigned any label annotated on this image with equal probability, but must not be assigned to those categories that are absent in this image. Here $|\{c_k\}|$ denotes the number of categories annotated for this image. This exclusive information is useful for reducing false positive predictions.

The second part of the unary term respects the output of a fully convolutional network. We denote this unary term as $\psi_i^{net}$, and define it as:

$$\psi_i^{net}(y_i) = -\log P(y_i|X, \Theta), \quad (3)$$

Here $\Theta$ represents the parameters of the network. $\log P(y_i|X, \Theta)$ denotes the log probability of predicting $x_i$ to have the label $y_i$. It is simply the summation of the pixel-wise log probability of all pixels in the super-pixel $x_i$. The two parts of the unary terms are illustrated in Fig. 2. The unary term $\psi_i$ is simply $\psi^{scr} + \psi^{net}$, for which an implicit balance weight of 1 is used.

The pairwise term $\psi_{ij}$ in Eqn.(1) models the similarity between two super-pixels. We only adopt a pairwise term to adjacent super-pixels. We consider simple appearance similarities for adjacent super-pixels. We build color and texture histograms for $x_i$. The color histogram $h_c(x_i)$ on $x_i$ is built on the RGB space using 25 bins for each channel. The texture histogram $h_t(x_i)$ is built on the gradients at the horizontal and the vertical orientations, where 10 bins are used for each orientation. All bins are concatenated and normalized in the color/texture histograms respectively. We define the pairwise term as:

$$\psi_{ij}(y_i, y_j|X) = [y_i \neq y_j] \exp\Big\{ -\frac{\|h_c(x_i) - h_c(x_j)\|_2^2}{\delta_c^2} - \frac{\|h_t(x_i) - h_t(x_j)\|_2^2}{\delta_t^2} \Big\}. \quad (4)$$

Here $[\cdot]$ is 1 if the argument is true and 0 otherwise. The parameters $\delta_c$ and $\delta_t$ are set as 5 and 10 respectively. This definition means that for the adjacent super-pixels assigned different labels, the cost is higher if their appearance is closer.

With these definitions, we have the optimization problem in this form:

$$\sum_i \psi_i^{scr}(y_i|X, S) + \sum_i -\log P(y_i|X, \Theta) + \sum_{i,j} \psi_{ij}(y_i, y_j|X), \quad (5)$$

where there are two sets of variables to be optimized: $Y = \{y_i\}$ for labeling all super-pixels, and $\Theta$ for the fully convolutional network's parameters. Eqn.(5) is for one training image. The total loss function sums over the loss functions for all training images.

## 2.2. Optimization

We present an alternating solution to optimize the above loss function. We fix $\Theta$ and solve for $Y$, and vice versa. The two alternating steps have clear intuitions: (i) with $\Theta$ fixed, the solver propagates labels to unmarked pixels, based on scribbles, appearance, and also network predictions; (ii) with $Y$ fixed, the solver learns a fully-convolutional network for pixel-wise semantic segmentation.

*Propagating scribble information to unmarked pixels.* With $\Theta$ fixed, the unary term of $\psi_i = \psi_i^{scr} + \psi_i^{net}$ can be easily evaluated by enumerating all possible labels $0 \leq y_i \leq C$. The pairwise terms can be pre-computed as a look-up table. With these evaluated unary and pairwise terms, the optimization problem in Eqn.(5) can be solved by the graph cuts solution [3, 2]. We use the multi-label graph cuts solver in

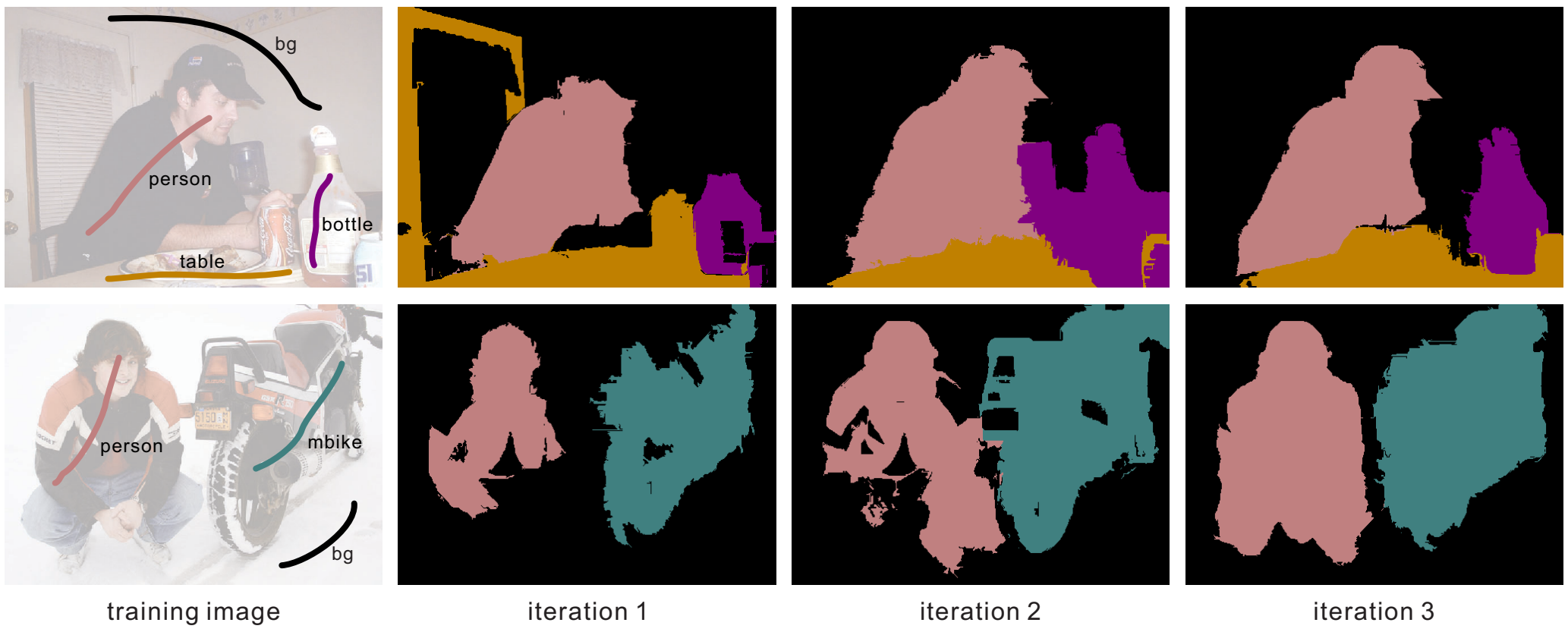

Figure 3. The propagated labels of a training image during the training process. Because of the network predictions that are aware of the high-level semantics, the propagated labels become more accurate.

[2] with publicly available code[2]. This solver assigns a category label $0 \leq y_i \leq C$ for each super-pixel $x_i$, and thus effectively propagate information to all unmarked pixels.

*Optimizing network parameters.* With the labeling $Y$ fixed for all super-pixels, solving for the network parameters $\Theta$ is equivalent to optimizing a network using the full-image annotation $Y$ as the supervision. In this paper, we adopt the FCN [21] as our network model. Given the the labeling $Y$, every pixel in the training image has been assigned a category label. So the FCN is directly trainable for handling this pixel-wise regression problem. The last layer of FCN outputs the per-pixel log probability, which are used to update the unary term in the graph.

The alternating solution is initialized from the graph cut step without using network prediction, and then iterates between the two steps. For every network optimizing step, we fine-tune the network with a learning rate of 0.0003 for 50k mini-batches and 0.0001 for the next 10k mini-batches, using a mini-batch size of 8. The network is (re-)initialized by an ImageNet [28] pre-trained model (*e.g.*, VGG-16 [30]) at the beginning of every network optimizing step. We have also experimented with going on training with the network parameters learned by the last iteration, and observed slightly poorer accuracy. We conjecture that this is because when the labels are not reliable, the network parameters might be tuned to a poorer local optimum.

Fig. 3 shows the evolution of the propagated labels on a training image. With the network being updated, the sematic information from the network becomes more reliable, and the propagated labels are getting more accurate. These propagated labels in turn improve network learning. We empirically find that the alternating algorithm converges after three iterations, and more iterations have negligible improvements.

[2]`http://vision.csd.uwo.ca/code/gco-v3.0.zip`

For inference, we only need to apply the FCN model (represented by $\Theta$) on the image. We note that the super-pixels and their graphical model are for training only, and are not needed for inference. Following [5], we adopt a CRF to post-process the results.

### 2.3. Related Work

*Graphical models for segmentation.* Graphical models are widely used for interactive image segmentation [4, 26, 17, 11, 16, 19] and semantic segmentation [15, 29, 13, 14]. Graphical models in general involves unary and pairwise terms, and are particularly useful for modeling local and global spatial constraints. Interestingly, FCN [21], as one of the recent most successful method for semantic segmentation, performs pixel-wise regression and thus only has explicit unary terms. But graphical models (*e.g.*, CRF/MRF) are later developed for FCN as post-processing [5] or joint training [31, 20]. The graphical models in [5, 31, 20] are developed for strongly-supervised training and their effects are mainly in improving mask boundaries [5], whereas our graphical model plays a more important role in our scenario for propagating information into the unknown pixels of the training images. The graphical models in [5, 31, 20] operates on pixels, in contrast to ours that operates on super-pixels. Pixel-based models are preferred for refining boundaries, but our super-pixel-based model can propagate labels more easily into regions that are distant from the scribbles.

*Weakly-supervised semantic segmentation.* There have been a series of methods [25, 24, 7, 27] on weakly-supervised learning CNN/FCN for semantic segmentation. *Image-level* annotations are more easily to obtain, but semantic segmentation accuracy [25, 24] by using only image-level labels lags far behind strongly-supervised results. Results based on *box-level* annotations [24, 7] are much closer to strongly-supervised ones. Alternating solvers are adopted in [24, 7] for addressing box-supervised training, as is also used by our method. But box annotations provide the object bounds and confident background regions, so it is not demanded for the box-supervised training to propagate information. Pairwise terms as ours are not considered in [24, 7], and graphical models are not used in their weakly-supervised training (except for CRF refinements). On the contrary, we will show by experiments that information propagation is influential in our scribble-supervised case.

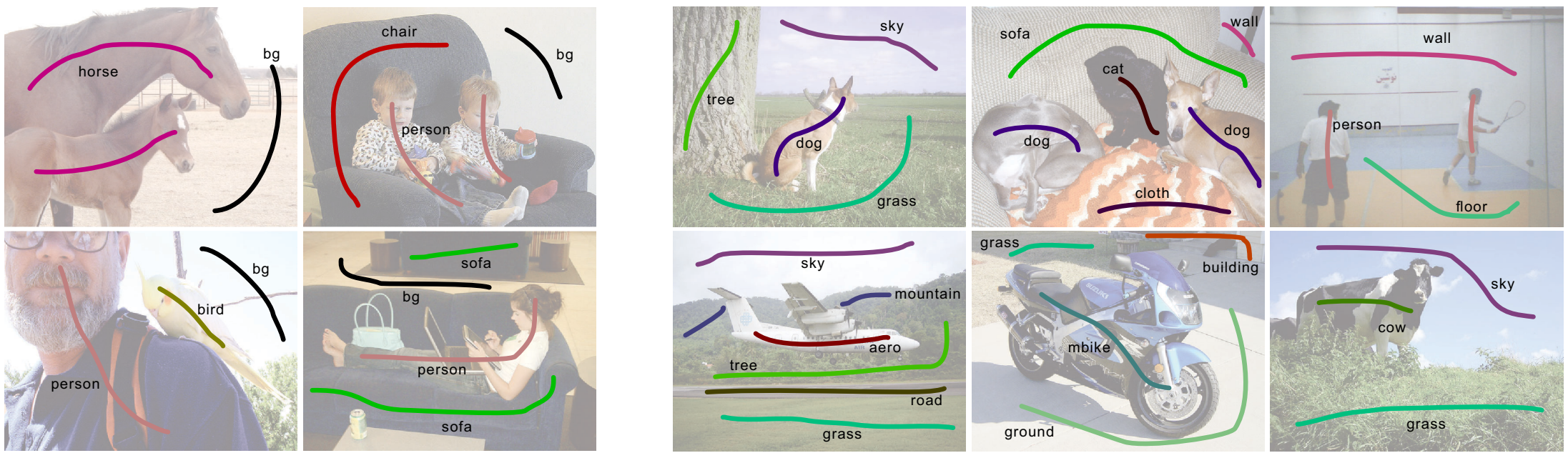

Figure 4. Examples of annotated scribbles via the crowdsourcing AMT platform. Left: PASCAL VOC 2012 that has 20 object categories. Right: PASCAL-CONTEXT that has 59 categories of objects and stuff.

| method | mIoU (%) |
|---|---|
| GrabCut+FCN | 49.1 |
| LazySnapping+FCN | 53.8 |
| ours, w/o pairwise terms | 60.5 |
| ours, w/ pairwise terms | **63.1** |

Table 1. Semantic segmentation results on the PASCAL VOC 2012 validation set via different strategies of utilizing scribbles.

# 3. Experiments

## 3.1. Annotating Scribbles

To evaluate our method, we use the Amazon Mechanical Turk (AMT) platform to obtain scribble annotations on the PASCAL VOC datasets. We annotate the PASCAL VOC 2012 set [9] that involves 20 object categories and one background category, and the PASCAL-CONTEXT dataset [22] that involves 59 categories of objects and stuff. We further annotate the PASCAL VOC 2007 set using the 59 categories (which include the 20 object categories). We note that the 2007 set has no available mask-level annotations. So although the scribble annotations for the 2012 and CONTEXT sets are mainly for the investigation purpose (as these sets have available masks), the scribble annotations for the 2007 set can be actually exploited to improve results.

Our scribble annotations were labeled by 10 annotators from the AMT platform. Each image is annotated by one annotator, and is checked by another annotator (if necessary, to add missing scribbles and modify imprecise scribbles). The annotators are asked to draw scribbles on the regions which they feel confident and easy to draw; the object boundaries or ambiguous regions are not needed to annotated. However, we require that every existing object (of the related categories) in an image must be labeled by at least one scribble, so missing objects are not allowed (verified by the checker annotator). According to our record, it takes an annotator on average 25 seconds to label an image with 20 object categories, and 50 seconds with 59 object/stuff categories. Compared with annotating per-pixel masks that takes several minutes per image [22, 18], the annotation effort by scribbles is substantially reduced. Fig. 4 shows some annotated scribbles. The scribbles have an average length of $\sim 70\%$ of the longer side of the object's bounding box.

## 3.2. Experiments on PASCAL VOC 2012

Following [21, 24, 7], we train the models on the 10,582 (denoted as 11k) training images [12] and evaluate on the 1,449 validation images. The accuracy is evaluated by the mean Intersection-over-Union (mIoU) score. We adopt the DeepLab-MSc-CRF-LargeFOV [5] as our strongly(mask)-supervised baseline (using VGG-16 [30]). Our implementation has 68.5% mIoU score, reasonably close to 68.7% reported in [5]. This network architecture also serves as our FCN structure in our scribble-supervised models.

**Strategies of utilizing scribbles**

Our method jointly propagates information into unmarked pixels and learns network parameters. A simpler solution is to first use any existing interactive image segmentation method to generate masks based on scribbles, and then use these masks to train FCNs. In Table 1 we compare with this two-step solution.

We investigate two popular interactive image segmentation algorithms for generating masks from scribbles: GrabCut[3] [26] and LazySnapping [17]. In our scenario, the difference between [26] and [17] lies on their definitions of the unary and pairwise terms. Training FCNs using the masks generated by these methods shows inferior semantic segmentation accuracy (Table 1, all post-processed by CRFs of [5]). This is because the terms in these traditional methods [26, 17] only model low-level color/sptial information and are unaware of semantic content. The generated masks are not reliable "ground-truth" for training the networks.

[3] Although GrabCut used boxes as it was originally developed, its definitions of unary/pairwise terms can be directly applied with scribbles.

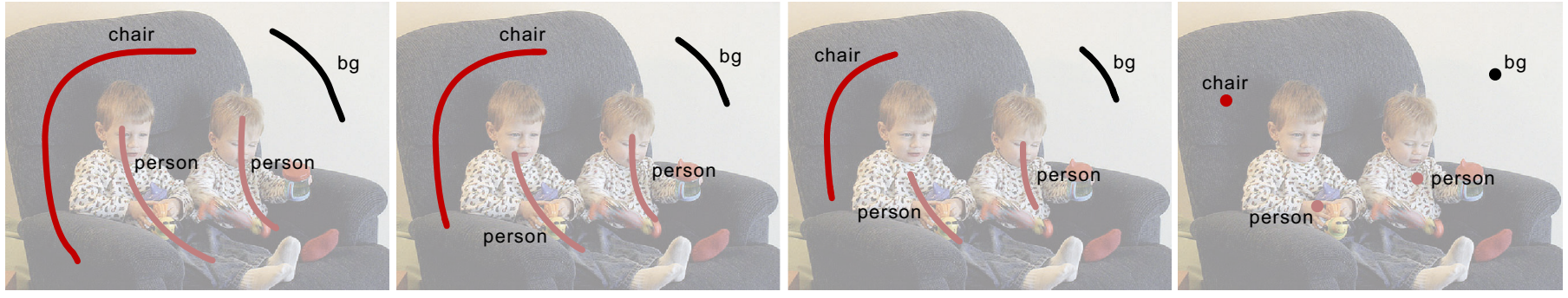

Figure 5. Scribbles of different lengths for investigating our method's sensitivities to scribble quality. See also Table 2.

| length ratio | mIoU (%) |
|---|---|
| 1 | 63.1 |
| 0.8 | 61.8 |
| 0.5 | 58.5 |
| 0.3 | 54.3 |
| 0 (spot) | 51.6 |

Table 2. Sensitivities to scribble length evaluated on the PASCAL VOC 2012 validation set. The shorter scribbles are synthesized from the annotated scribbles, reducing their length by a ratio. A ratio of 0 means spots are used.

On the contrary, our scribble-supervised method achieves a score of 63.1%, about 10% higher than the two-step solutions. This is because when the network gradually learns semantic content, the high-level information can help with the graph-based scribble propagation. This behavior is shown in Fig. 3: at the beginning of training when the network is not reliable, the propagated labels tend to rely on low-level color/texture similarities; but the erroneous propagation can be corrected with the network feedback.

In Table 1 we also show that the pairwise term is important for propagating information. We experiment with a case without using the pairwise terms, which leads to a lower score of 60.5%. In this case, the unmarked pixels are ambiguous at initialization (as their cost of taking any available label is equal), so we initialize them by the background label. Without the pairwise term, the problem in Eqn.(5) only involves unary terms, and the graph cuts step actually degrades to winner-take-all selection of labels based on network predictions. As such, the information propagation is only performed by the "fully convolutional" behavior (sliding a large receptive field) that may implicitly impose local coherency. But this way of propagation is insufficient as shown by the lower accuracy.

**Sensitivities to scribble quality**

The quality of the annotated scribbles is subject to the behavior and experience of the annotators. So we investigate how sensitive our method is to the scribble quality. Because the "quality" of scribbles is somehow subjective and using different sets of scribbles requires extra annotation efforts, we use synthesized scribbles based on the user annotations. We focus on the length of the scribbles. Given any user annotated scribble, we reduce the length of this scribble by a ratio. For a shortened scribble, one of its end-points is randomly chosen from the original scribble, and the other end-point is determined by the reduced length. Fig. 5 shows examples of the shortened scribbles, where a ratio of 0 means a spot is used.

| method | annotations | mIoU (%) |
|---|---|---|
| MIL-FCN [25] | image-level | 25.1 |
| WSSL [24] | image-level | 38.2 |
| point supervision [27] | spot | 46.1 |
| WSSL [24] | box | 60.6 |
| BoxSup [7] | box | 62.0 |
| ours | spot | 51.6 |
| ours | scribble | **63.1** |

Table 3. Comparisons of weakly-supervised methods on the PASCAL VOC 2012 validation set, using different ways of annotations. All methods are trained on the PASCAL VOC 2012 training images using VGG-16, except that the annotations are different.

Table 2 shows the results of our scribble-supervised algorithm using scribbles of different lengths. Our method performs gracefully when the scribbles are shortened, suggesting that our method is reasonably robust to the quality of scribbles. To the extreme, when the length approaches 0 and the scribbles become spots, our method is still applicable and has a score of 51.6%.

**Comparisons with other weakly-supervised methods**

In Table 3 we compare with other weakly-supervised methods using different ways of annotations. We note that while image-level annotations are the most economical, their weakly-supervised learning accuracy (*e.g.*, 38.2% of WSSL [24]) lags far behind other ways of annotations. On the other hand, our method achieves accuracy on par with box-supervised methods (60.6% of of WSSL [24] and 62.0% of BoxSup [7]), indicating that scribbles can be well exploited as a user-friendly alternative to boxes. Comparing with a recent point-supervised method [27] (46.1%), our spot-only result (51.6%) is over 5% higher (but we note that the our annotations are different from those used in [27]).

| method | data/annotations | mIoU (%) |
|---|---|---|
| CFM [8] | 5k w/ masks | 34.4 |
| FCN [21] | 5k w/ masks | 35.1 |
| Boxsup [7] | 5k w/ masks + 133k w/ boxes (COCO+VOC07) | 40.5 |
| baseline | 5k w/ masks | 37.7 |
| ours, weakly | 5k w/ scribbles | 36.1 |
| ours, weakly | 5k w/ scribbles + 10k w/ scribbles (VOC07) | 39.3 |
| ours, semi | 5k w/ masks + 10k w/ scribbles (VOC07) | **42.0** |

Table 5. Comparisons on the PASCAL-CONTEXT validation set.

| supervision | # w/ masks | # w/ scribbles | total | mIoU (%) |
|---|---|---|---|---|
| weakly | - | 11k | 11k | 63.1 |
| strongly | 11k | - | 11k | 68.5 |
| semi | 11k | 10k (VOC07) | 21k | **71.3** |

Table 4. Comparisons of our method using different annotations on the PASCAL VOC 2012 validation set. The term "total" shows the number of training images, "# w/ masks" shows the number of training images with mask-level annotations, and "# w/ scribbles" shows the number of training images with scribble annotations.

**Comparisons with using masks**

In Table 4 we compare our weakly-supervised results based on scribbles and strongly-supervised results based on masks. When replacing all mask-level annotations with scribbles, our method has a degradation of about 5 points. We believe this is a reasonable gap considering the challenges of exploiting scribbles.

Our method can also be easily generalized to semi-supervised learning that uses both mask annotations and scribble annotations. For the training images that have mask-level annotations, the graphical model is not applied, and these images only contribute to the network training step. To evaluate our semi-supervised results, we use the mask-level annotations on the PASCAL VOC 2012 set, and the scribble annotations on the PASCAL VOC 2007 set (that has no available mask annotation). Our semi-supervised result is a score of 71.3%, showing a gain of 2.8% higher than the baseline. This gain is due to the extra scribble annotations from the 2007 set. As a comparison, [24] reports a strongly-supervised result of 71.7% on this validation set using extra 123k COCO [18] images with masks. Our semi-supervised result is on par with their strongly-supervised result, but we only use 10k VOC 2007 scribble-level annotations as the extra data. Fig. 6 shows some results.

We further evaluate our method on the PASCAL VOC 2012 *test* set. By semi-supervised training using the mask-level annotations of the PASCAL VOC 2012 train and validation sets, as well as the scribble annotations on the VOC 2007 set, the trained model has a score of 73.1%. This number is behind but close to the current state-of-the-art results on the test set[4], without using the 123k COCO data. The competitive accuracy of our method suggests that using inexpensive scribbles to increase the data size can be a practical solution.

### 3.3. Experiments on PASCAL-CONTEXT

We perform more experiments on the PASCAL-CONTEXT dataset [22] with 59 categories. The images of this dataset are fully annotated by [22], providing pixel-level masks on objects and stuff. We evaluate on the 5k validation images, and train the models on the 5k training images or extra data. We note that scribbles are particularly favored for annotating stuff, and annotating precise outlines of stuff can be more time-consuming than objects. Table 5 compares the results.

Our reproduced strongly-supervised baseline has a score of 37.7% on this dataset. Our weakly-supervised method based on scribbles has a score of 36.1%, showing a graceful degradation of 1.6 point. This gap is smaller than that for PASCAL VOC 2012 (Table 4), suggesting that it is easier to propagate stuff labels as stuff in general has uniform appearance. When using the extra (object+stuff) scribbles annotated on the PASCAL VOC 2007 set, our weakly-supervised result is boosted to 39.3%, demonstrating the effects of exploiting more training data (which are yet efficient to annotate). Finally, our semi-supervised method that uses the provided 5k mask-level annotations and extra 10k VOC 2007 scribble-level annotations achieves a score of 42.0%. This number compares favorably with BoxSup [7] (40.5%) that exploits extra 133k box-level annotations ($\times 10$ of ours). We note that box-level annotations are only applicable for objects, but is not for stuff. So even though BoxSup uses much larger amount of data than ours, its accuracy is lower than ours on this dataset that involves stuff. To the best of our knowledge, our accuracy is the current state of the art on this dataset. Fig. 7 shows some example results.

## 4. Conclusion and Future Work

We have presented a weakly-supervised method for semantic segmentation based on scribbles. Our method opti-mizes a graphical model for propagating information from scribbles. Although our formulation is motivated by the usage of scribbles, it is applicable for many other types of weak supervision including box-level or image-level annotations. We plan to investigate these issues in the future.

[4]Using only the VOC data, the strongly-supervised methods of [23, 20] have test set accuracy of over 74.8% and 74.1%. The improvements of [23, 20] are orthogonal to our method, and these strongly-supervised methods can be swapped into our system, in place of the FCN baseline.

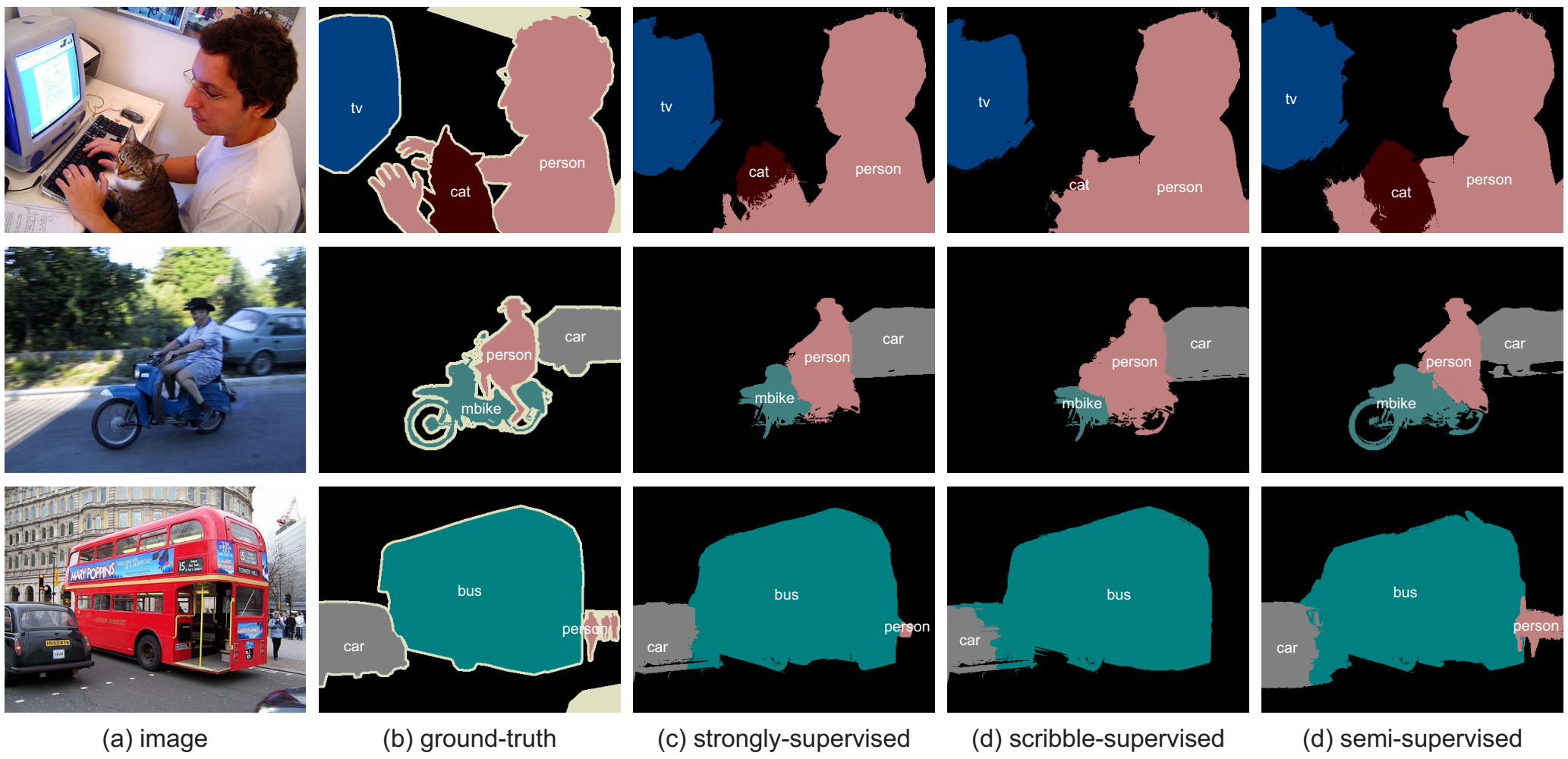

Figure 6. Our results on the PASCAL VOC 2012 validation set. The training data and annotations are: (c) 11k images with masks on VOC 2012; (d) 11k images with scribbles on VOC 2012; (e) 11k images with masks on VOC 2012 and 10k images with scribbles on VOC 2007. See also Table 4.

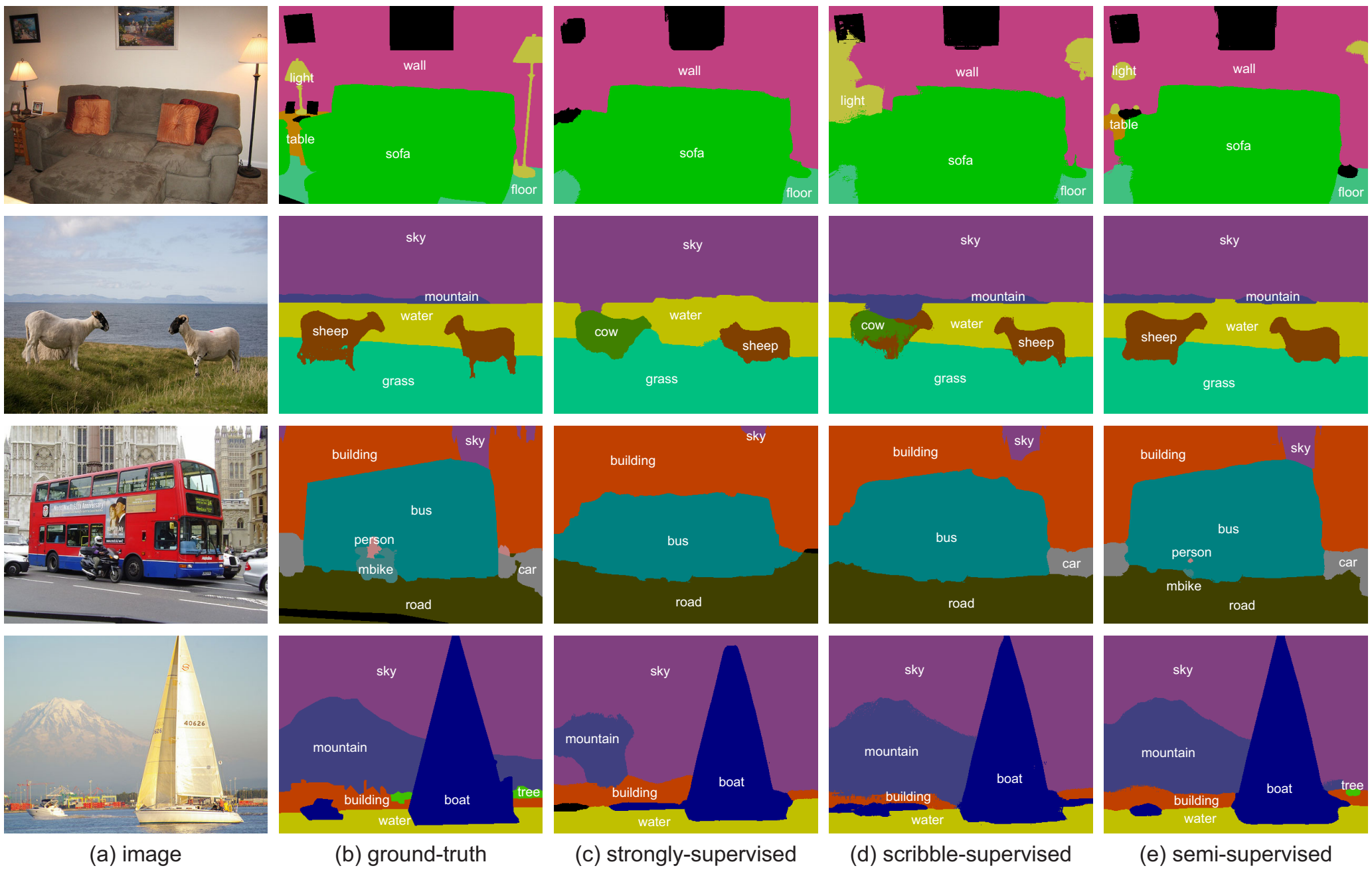

Figure 7. Our results on the PASCAL-CONTEXT validation set. The training data and annotations are: (c) 5k images with masks on PASCAL-CONTEXT; (d) 5k images with scribbles on PASCAL-CONTEXT; (e) 5k images with masks on PASCAL-CONTEXT and 10k images with scribbles on VOC 2007. See also Table 5.

## Acknowledgment

This work is in part supported by a grant from the Research Grants Council of the Hong Kong SAR (project No. 413113).